\documentclass[runningheads]{llncs}

\usepackage{eccv}

\usepackage{eccvabbrv}

\usepackage{graphicx}
\usepackage{booktabs}

\usepackage[accsupp]{axessibility}  

\usepackage[pagebackref,breaklinks,colorlinks,citecolor=eccvblue]{hyperref}

\usepackage{orcidlink}

\usepackage{csquotes}   
\usepackage{multirow}
\usepackage{booktabs}   
\usepackage{siunitx}    
\usepackage{graphicx}   
\usepackage{xcolor}     
\usepackage{colortbl}   
\usepackage{diagbox}    
\usepackage{amsmath}    

\definecolor{sectionblue}{RGB}{225, 240, 245}
\definecolor{sectiontext}{RGB}{50, 50, 50} 
\definecolor{rankonegreen}{RGB}{220, 240, 220}
\definecolor{ceiling}{RGB}{214,  38, 40}   
\definecolor{floor}{RGB}{43, 160, 4}     
\definecolor{wall}{RGB}{158, 216, 229}  
\definecolor{window}{RGB}{114, 158, 206}  
\definecolor{chair}{RGB}{204, 204, 91}   
\definecolor{bed}{RGB}{255, 186, 119}  
\definecolor{sofa}{RGB}{147, 102, 188}  
\definecolor{table}{RGB}{30, 119, 181}   
\definecolor{tvs}{RGB}{160, 188, 33}   
\definecolor{furniture}{RGB}{255, 127, 12}  
\definecolor{objects}{RGB}{196, 175, 214} 

\definecolor{geoWeakBlue}{HTML}{E3F2FD}  
\definecolor{geoStrongBlue}{HTML}{90CAF9} 

\definecolor{semWeakOrange}{HTML}{FFF3E0} 
\definecolor{semStrongOrange}{HTML}{FFB74D} 

\definecolor{textBlue}{HTML}{1565C0}   
\definecolor{textOrange}{HTML}{E65100} 

\begin{document}

\title{YouTube-Occ: Learning Indoor 3D Semantic Occupancy Prediction from YouTube Videos} 

\titlerunning{YouTube-Occ}

\author{Haoming Chen\inst{1}\orcidlink{0000-0002-4137-0417} \and
Lichen Yuan\inst{1} \and 
Tianfang Sun\inst{1}\orcidlink{0000-0003-2536-3302} \and
Jingyu Gong\inst{1,2,3}\textsuperscript{\dag}\orcidlink{0000-0002-4536-0953} \and
\\
Zhizhong Zhang\inst{1,3}\orcidlink{0000-0001-6905-4478} \and
Xin Tan\inst{1,2}\orcidlink{0000-0001-9346-1196}  \and
Yanyun Qu\inst{4}\orcidlink{0000-0002-8926-4162} \and
Yuan Xie\inst{1,2}\textsuperscript{\dag}\orcidlink{0000-0001-6945-7437}
}

\authorrunning{H.~Chen et al.}

\institute{School of Computer Science and Technology, East China Normal University, China \and
Chongqing Key Laboratory of Precision Optics, Chongqing Institute of East China Normal University, China \and
Shanghai Key Laboratory of Computer Software Evaluating and Testing, China\and
School of Informatics, Xiamen University, China
}

\maketitle
\renewcommand{\thefootnote}{\dag}
\footnotetext{Corresponding author}
\renewcommand{\thefootnote}{\arabic{footnote}}

\begin{abstract}
3D semantic occupancy prediction is crucial for fine-grained scene understanding, yet its advancement in  privacy-sensitive indoor environments is fundamentally hindered by the scarcity of large-scale annotated 3D data. To overcome this limitation, we explore learning indoor 3D semantic occupancy prediction from abundant, uncalibrated in-the-wild internet videos while simultaneously bypassing the extensive manual annotation. Specifically, we introduce \textit{YouTube-Occ}, including an automated data pipeline that leverages 2D and 3D foundation models to process raw web videos, estimating camera geometry, reconstructing scene point clouds, and enriching them with dense semantic pseudo-labels. However, these plausible pseudo-labels fail to yield performance gains under naive supervision. To address this impasse, we further propose a pre-training framework driven by feature distillation with a dual-alignment strategy. Within it, an intra-frame alignment utilizes a voxel-anchored Gaussianization module to align 3D features with corresponding 2D priors, whereas a cross-scene alignment achieves global semantic consistency via  class-prototype distillation. Empirically, YouTube-Occ delivers consistent gains across three mainstream architectures on the NYUv2 and Occ-ScanNet benchmarks, especially under limited-data conditions. We will publicly release our code and data, hoping to inspire future research.


\end{abstract}    
\section{Introduction}
\label{sec:intro}

\begin{figure}[htbp]
\centering
\includegraphics[width=0.95\linewidth]{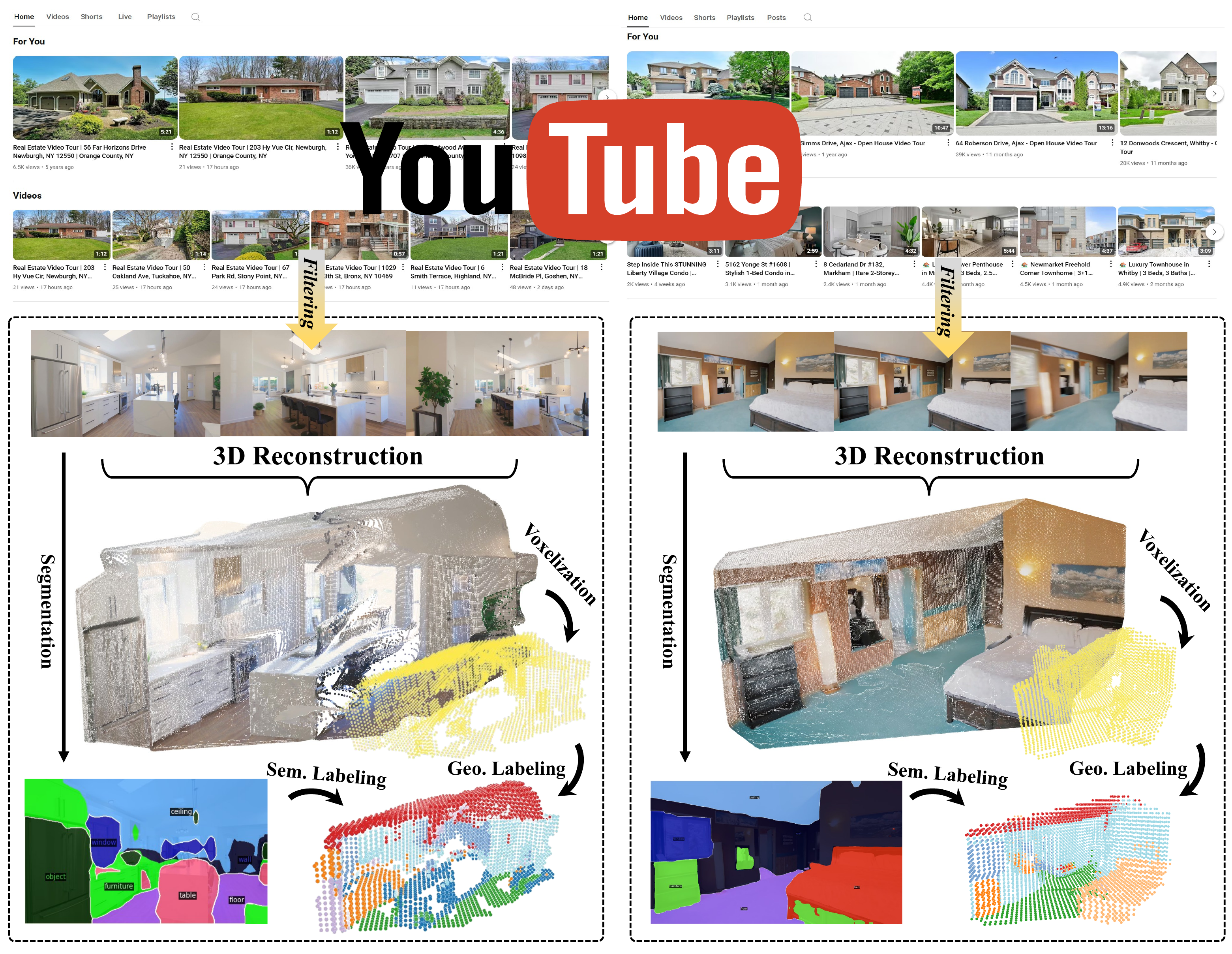}
\caption{YouTube-Occ could acquire diverse and rich indoor videos from the YouTube platform and support 3D semantic occupancy training without pre-configured camera.}
\label{fig:motivation}
\vspace{-2em}
\end{figure}

As one of the most promising abilities of artificial intelligence, 3D scene understanding is becoming increasingly important for applications such as autonomous driving~\cite{caesar2020nuscenes} and robotic navigation~\cite{shah2023lm}. Just as humans possess a natural ability to comprehend 3D environments through vision, 3D scene perception plays a fundamental role in understanding surroundings for machines. Among them, semantic occupancy prediction~\cite{cao2022monoscene, jiang2024symphonize} discretizes 3D space into voxel grids with semantic labels, which provide a fine-grained and unified geometric representation, leading to a detailed perception of object shapes and scene-level structures.

However, the advancement of this field, especially for privacy-sensitive indoor environments, is fundamentally bottlenecked by the scarcity of large-scale and annotated 3D data. Indoor scenes present formidable challenges due to intricate geometries, severe occlusions, and wide variations in object scales~\cite{yu2024monocular}. Existing indoor semantic occupancy benchmarks are limited in scale and diversity (\emph{e.g.}, NYUv2~\cite{silberman2012indoor} is only 1,449 images). Moreover, the manual collection and labeling for indoor scene is labor-intensive and unscalable. The high cost of specialized 3D scanning equipment, compounded by significant privacy concerns, also creates a critical barrier. To bypass these barriers, we ask a compelling question: \textit{Can we learn 3D indoor semantic occupancy from abundant, unlabeled, and uncalibrated in-the-wild internet videos?}

To answer this question, we propose \textbf{YouTube-Occ}, which incorporates an automatic data pipeline capable of producing geometric and semantic pseudo-labels from in-the-wild internet videos. As illustrated in Fig.~\ref{fig:motivation}, our pipeline curates video clips and leverages modern 3D foundation models to jointly estimate camera geometry and reconstruct scene point clouds. These point clouds are then enriched with dense semantic labels derived from 2D foundation models, culminating in a rich data source for pre-training existing 3D semantic occupancy models.

While utilizing foundation models to generate pseudo-labels seems intuitive, our empirical analysis indicates that direct supervision with pseudo-labels yields unsatisfactory results. Inspired by advancements in outdoor image-to-LiDAR distillation~\cite{sautier2022image,chen2024building,xu20244d,zou2024unim}, we design a  pre-training framework tailored for both in-domain and cross-domain fine-tuning. Crucially, this framework empowers existing semantic occupancy models to effectively harness unconstrained internet videos. Specifically, the core of our framework is a dual alignment strategy driven by two complementary contrastive objectives. First, an intra-frame feature alignment mechanism utilizes a voxel-anchored Gaussianization module to render 3D features, matching 2D priors from the corresponding view. Second, a cross-scene feature alignment module enforces global semantic consistency by aligning representations with dynamically updated class prototypes. Extensive evaluations on NYUv2 and Occ-ScanNet show consistent performance gains across three semantic occupancy architectures. Moreover, leveraging internet videos could enhance model performance in data-scarce scenarios.

\begin{itemize}
    \item To address data scarcity and annotation bottlenecks in indoor semantic occupancy prediction, we propose an automated data pipeline that, to our knowledge, is the first to use unlabeled, uncalibrated in-the-wild videos for 3D semantic occupancy prediction.
    \item  We introduce a  pre-training framework compatible with mainstream semantic occupancy architectures. Using voxel-anchored Gaussian rendering and class-prototype distillation, it learns generalizable representations in both intra- and cross-domain settings.
    \item Extensive experiments demonstrate that our approach reliably boosts performance on three popular indoor semantic occupancy architectures, with particularly substantial improvements in few-shot scenarios.
\end{itemize}

\section{Related Work}
\vspace{-0.5em}
\subsection{3D Semantic Occupancy}
\vspace{-0.5em}
3D semantic occupancy prediction, which represents scenes as fine-grained semantic grids, has become a pivotal task for holistic 3D understanding. A significant body of prior work \cite{cao2022monoscene, huang2021bevdet, li2022bevformer, Yang2022BEVFormerVA, huang2023tri, wei2023surroundocc, zhang2023occformer, li2023voxformer, li2023fbocc, ma2024cotr} has focused on advancing performance through architectural innovations, such as sophisticated feature fusion and temporal modeling~\cite{chen2025occprophet, li2024viewformer, zheng2023occworld, yang2024driving}, or by integrating high-level reasoning with large language models~\cite{wei2024occllama, xu2025occ}. While these data-driven methods have achieved impressive results, they are constrained by large-scale, manually annotated 3D datasets. The prohibitive cost and privacy concerns associated with acquiring such data severely bottleneck the development of these models. Our work deviates from this model-centric paradigm by proposing a new data-centric approach to mitigate this core limitation.
\vspace{-0.5em}
\subsection{3D Self-Supervised Learning}
\vspace{-0.5em}
3D self-supervised learning (SSL) offers a compelling paradigm to reduce reliance on expensive manual annotations. A common approach involves learning representations by enforcing consistency between a 3D model and its 2D renderings~\cite{hayler2023s4c, huang2024selfocc, chubin2023occnerf}, often employing differentiable renderers like Neural Radiance Fields~\cite{mildenhall2020nerf} or 3D Gaussian Splatting~\cite{kerbl3Dgaussians}. However, these methods are predominantly tailored for multi-view outdoor scenes and presume the availability of pre-calibrated sensor data. In contrast, our approach is specifically designed to tackle the unique challenges of cluttered and varied indoor environments by leveraging uncalibrated in-the-wild videos. Our pre-training framework is introduced to effectively learn from this challenging data, addressing the object complexity and data scarcity inherent to indoor scenes.

\vspace{-0.5em}
\subsection{Web-data Learning}
\vspace{-0.5em}
Web-data learning offers a scalable and cost-effective alternative to creating datasets that involve substantial financial and operational burdens. Many methods have designed sophisticated data-cleaning pipelines to harness this data for various tasks~\cite{cherti2023reproducible, kong2024hunyuanvideo, oquab2023dinov2, schuhmann2021laion, schuhmann2022laion, gong2021omni,gong2021boundary,liu2021deep,feng2025high}. For instance, YouTube-VLN ~\cite{lin2023ytbvln} demonstrated success by collecting large-scale indoor video frames from YouTube for Vision-and-Language Navigation. However, the application of web-data learning to 3D semantic occupancy prediction remains an unexplored area. To our knowledge, our work is the first to bridge this gap. We present a complete pipeline that automatically curates, reconstructs, and pseudo-labels unstructured internet videos to create a promising pre-training dataset for 3D occupancy models.

\section{Automated Internet Data Curation}
We introduce an automated data pipeline to harvest and process massive, uncalibrated indoor videos (\emph{e.g.}, unknown camera information.) from the internet for 3D semantic occupancy prediction. First, we revisit the manual indoor occupancy annotation process in Sec. \hyperref[sec:data_revisiting]{3.1}. Then, we detail our data curation pipeline, which leverages 2D/3D foundation models to generate pseudo-labels in Sec. \hyperref[sec:data_pipeline]{3.2}. Finally, we discuss the challenges involved in using pseudo-labels in Sec. \hyperref[sec:how_to_use_pseudo_label]{3.3}.

\vspace{-1em}
\subsection{Revisiting Manual Indoor Occupancy Annotation}
\label{sec:data_revisiting}
In this section, we briefly summarize how indoor semantic occupancy labels are generated in Occ-ScanNet \cite{yu2024monocular}. Occ-ScanNet uses RGB images, depth maps, camera intrinsics, and camera poses from ScanNet \cite{dai2017scannet}, together with scene-level semantic voxel annotations from CompleteScanNet \cite{wu2020scfusion}. To obtain frame-level semantic occupancy labels, it defines a voxel origin from the current camera pose, specifying a 3D volume in front of the camera. This involves two main components: (1) \textbf{Scene Data Collection}: ScanNet \cite{dai2017scannet} provides RGB-D video of indoor scenes captured by handheld depth sensors, with accurate depth calibration and server-side 6-DoF camera trajectories; (2) \textbf{Scene-Level Semantic Voxels}: CompleteScanNet \cite{wu2020scfusion} combines 3D CAD models from ShapeNet \cite{wu20153dshapnet} with CAD-to-scan alignments from Scan2CAD \cite{avetisyan2019scan2cad} to produce scene-level semantic voxel grids encoding complete object geometry.

To alleviate the labor-intensive and prohibitively expensive burden of manual data collection and annotation, we propose a scalable paradigm that harvests indoor videos from the online video platform (\emph{e.g.,} YouTube). By leveraging off-the-shelf 2D and 3D foundation models, we automatically generate pseudo-labels at scale. These pseudo-labels will be used in our pre-training framework to enable efficient pre-training of different semantic occupancy models.

\subsection{Automated Data Curation Pipeline}
\label{sec:data_pipeline}
\begin{figure}[htbp]
\centering
\includegraphics[width=0.9\linewidth]{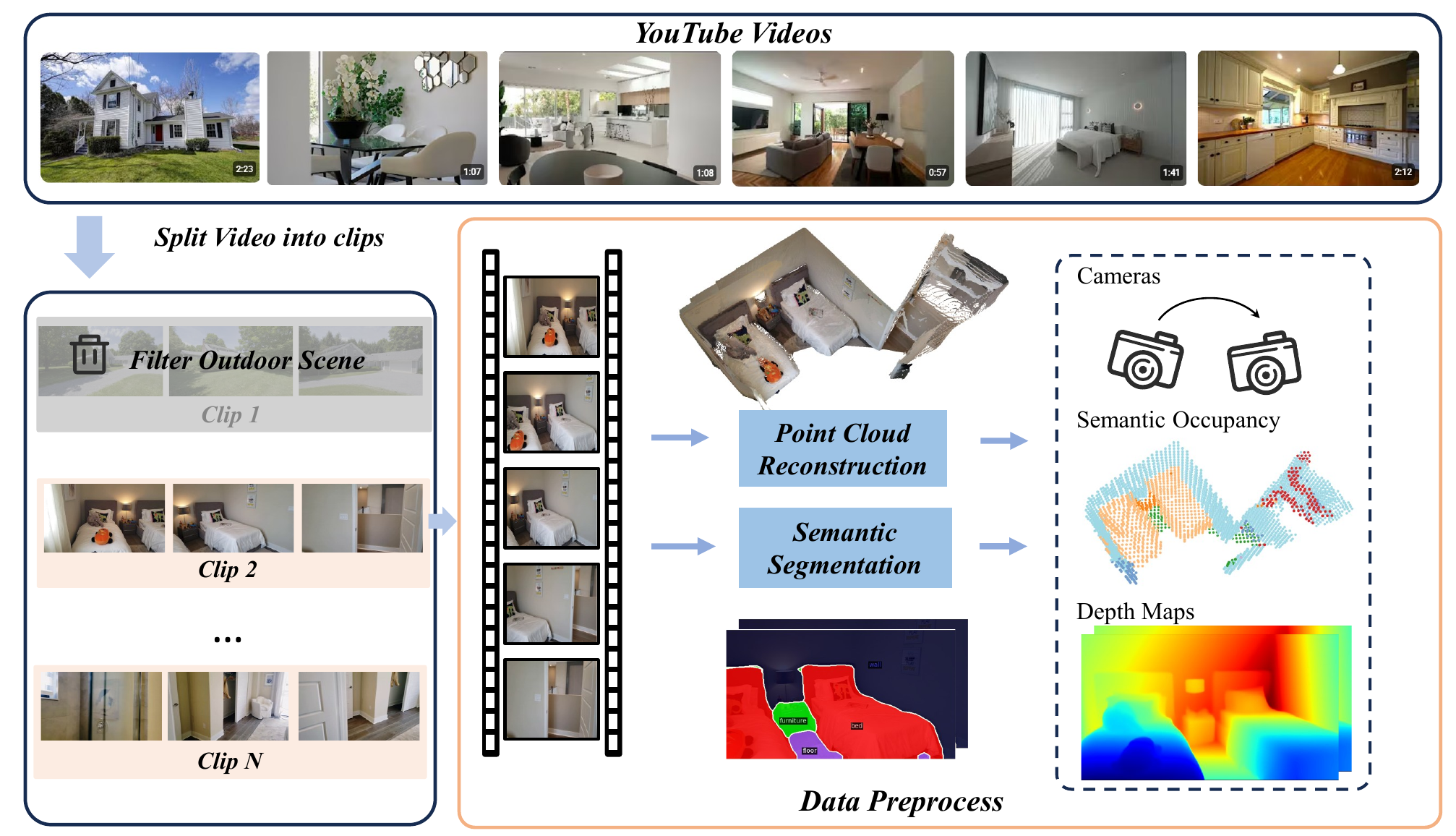}
\caption{ The data pipeline for generating pseudo-labels from internet videos. Starting from YouTube videos, we first split them into clips and filter out the irrelevant ones. We then use a 3D foundation model to reconstruct the surface point clouds of scene objects, and a 2D foundation model to extract pixel-level semantic information. Finally, each frame is associated with pseudo-labels, including camera intrinsics, camera poses, depth maps, 2D semantic maps, and 3D semantic occupancy grids.}
\label{fig:internet}
\end{figure}

The data pipeline, as in Fig.~\ref{fig:internet}, consists of video collection and filtering, geometry reconstruction, semantic segmentation, and semantic occupancy generation.

\textbf{Video Collection.} \quad Inspired by \cite{zhou2018stereo}, we choose the category of real estate footage as our video data source. In general, these real estate videos mainly contain indoor footage, which tend to have a smooth and steady camera motion. These shots mainly provide three major advantages for our data construction: 1) Easier 3D Reconstruction: Stable inter-frame motion helps reconstruction and reduces errors from motion blur and camera jitter. 2) Broader Scene Diversity: Internet videos span many environments, providing richer data that improves 3D perception model generalization. 3) Mostly Static Content: Beyond camera ego-motion, these videos are largely static, making them analogous to indoor benchmarks Occ-ScanNet. Specifically, we begin by collecting video IDs from real estate channels on the YouTube website and then use the video download tool yt-dlp to store these videos locally. Next, we split each video into separate clips following the procedure in \cite{zhou2018stereo}. To discard irrelevant segments, we utilize the off-the-shelf image classifier from \cite{zhou2017places} to filter out outdoor scenes.

\textbf{Point Cloud Reconstruction.} \quad Since in-the-wild internet videos provide only uncalibrated RGB images without camera parameters, we cannot directly apply existing 3D occupancy models. To reconstruct the 3D geometry, we employ Dust3R~\cite{wang2024dust3r} to jointly estimate camera intrinsics, poses, and point clouds. To address misalignments and scale ambiguity in the raw output, we apply a post-processing pipeline: we use heuristic filters to clean the data, transform the scene to a z-up orientation parallel to the floor, and enforce an approximate metric scale assuming a 2.8-meter wall height. Our design choices are driven by two considerations. (1) We adopt Dust3R~\cite{wang2024dust3r} as it marks the genesis of the recent boom in 3D foundation models (\emph{e.g.}, VGGT~\cite{wang2025vggt} and Pi3~\cite{wang2025pi}). (2) Since estimating exact metric scale from uncalibrated web data lacking any camera priors is an inherently ill-posed problem, the 2.8-meter architectural prior crucially normalizes diverse scenes into a consistent canonical metric space, ensuring stable occupancy voxelization at a fixed resolution of 0.08 m.


\textbf{Semantic Segmentation.} \quad Following \cite{chen2024building}, we utilize SAN \cite{xu2023side} for open-vocabulary semantic segmentation. The segmentation masks are obtained by directly inputting the textual label of each category from the indoor occupancy benchmarks (both NYUv2~\cite{silberman2012indoor} and Occ-ScanNet~\cite{yu2024monocular} contain 11 semantic classes) into the model. We deliberately eschew elaborate and transient prompt engineering to mitigate the risk of overfitting to specific formulations. This design choice guarantees a strictly automated pipeline, eliminating the need for task-specific human intervention while ensuring effortless scaling and future extensibility in complex open-world environments.

\textbf{Occupancy Generation.}\quad Given the point cloud and pixel-level semantics, we assign each point both spatial and semantic attributes. We next voxelize the semantic point cloud using a fixed voxel size (\emph{e.g.}, 0.08 m) to form semantic scene voxels, assigning each voxel the most common class label of its contained points. Finally, to derive frame semantic voxels, we specify a 3D bounding box over the front  area and label each voxel within it as either occupied or empty.

\vspace{-1em}
\subsection{Exploiting Pseudo-Labels}
\label{sec:how_to_use_pseudo_label}
A naive way to use our auto-labeled internet videos is to directly train existing 3D semantic occupancy networks (\emph{e.g.,} Symphonies\cite{jiang2024symphonize}) with these pseudo-labels under standard supervision. However, empirical results (Sec.~\hyperref[sec:evaluation_of_data_pipeline]{5.2}) show that direct  zero-shot inference  or joint training  leads to marginal or even negative transfer on benchmarks like Occ-ScanNet. Based on qualitative visualizations and quantitative analyses, we conjecture the reason is two misalignments in pseudo-labels: (1) Geometric scale misalignment: Monocular internet videos lack absolute metric scale, causing 3D foundation model predictions to be systematically mis-scaled relative to the real-world metric spaces of standard indoor datasets. (2) Semantic misalignment: 2D foundation model predictions are inconsistent with occupancy benchmark categories, lacking a strict one-to-one mapping to the target dataset taxonomy.

The inherent lack of geometric constraints and the high scene diversity in in-the-wild internet videos make direct supervision with unscaled pseudo-labels sub-optimal. To address this, we propose a pre-training framework that leverages such imperfect labels through voxel-anchored Gaussianization and class-prototype-based distillation. As detailed in Sec.~\ref{sec:evaluation_of_pretraining}, this approach yields consistent performance gains across three mainstream architectures and two semantic occupancy benchmarks, demonstrating its robustness to the noisy nature of web-sourced data.

\section{Pre-training Framework}
Our pre-training framework utilizes self-supervised distillation to enhance popular 3D semantic occupancy networks. First, we describe the background of a vision-centric occupancy prediction network in Sec. \hyperref[sec:revisiting]{4.1}. Then, we show the details of our pre-training method in Sec. \hyperref[sec:pre-training-stage]{4.2}. Finally, we discuss the differences between our method and related outdoor self-supervised paradigms in Sec.\ \hyperref[sec:pre-training-discussion]{4.3}.
\begin{figure}[htbp]
\centering
\includegraphics[width=0.9\linewidth]{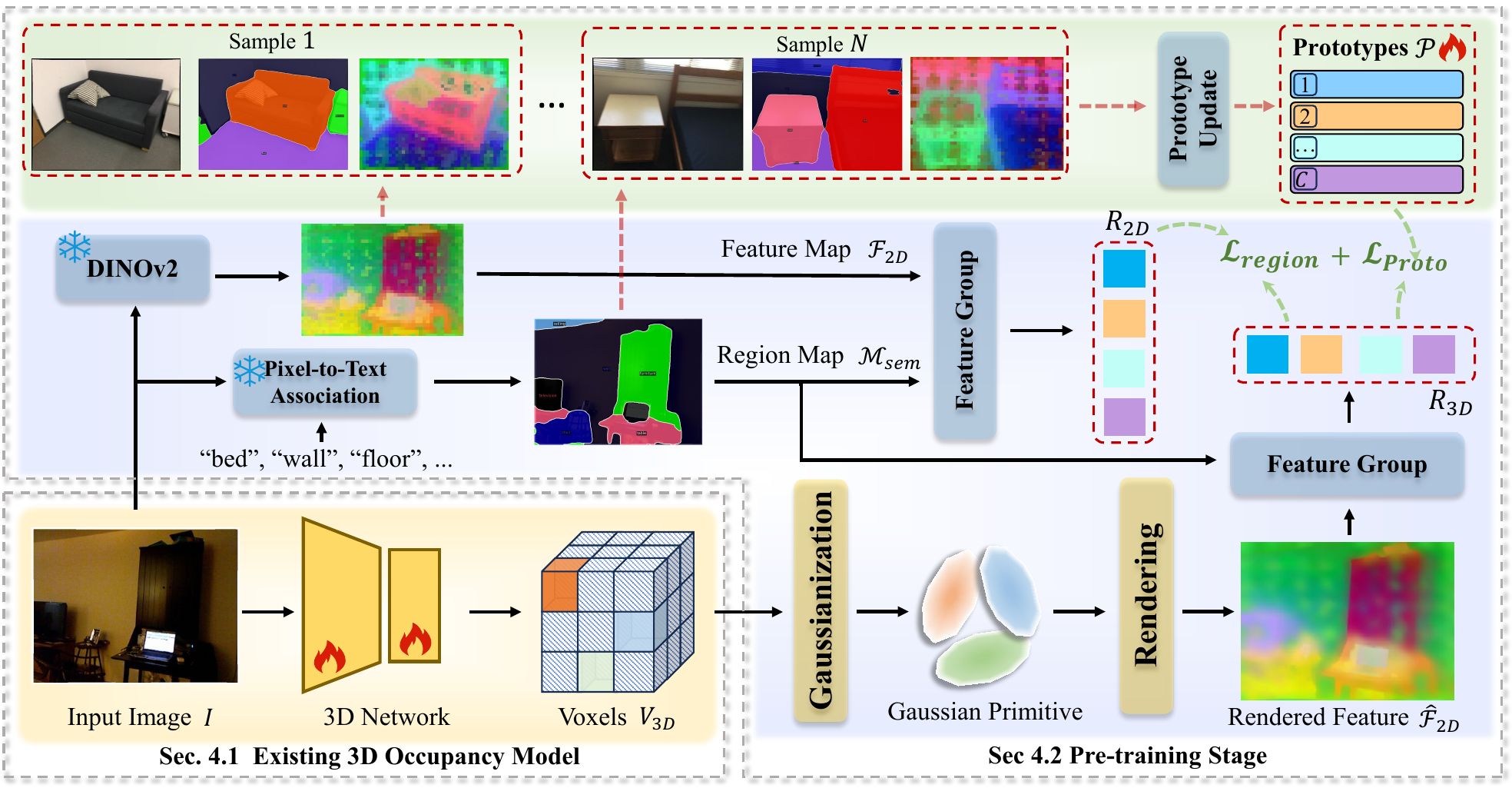}
\caption{ 
Overview of our pre-training framework. It supports distilling foundation model knowledge into mainstream 3D occupancy networks. To bridge the 3D-to-2D modality gap, we propose a Voxel-Anchored Gaussianization module to project 3D voxel features onto the 2D image plane. We design two complementary alignment objectives: an intra-frame region loss ($\mathcal{L}_{region}$) for local semantic coherence  and a cross-scene prototype loss ($\mathcal{L}_{proto}$) for global semantic consistency.
}
\label{fig:pre-training}
\end{figure}

\subsection{Revisiting Occupancy Prediction}
\label{sec:revisiting}
Classically, a vision-centric semantic occupancy model takes a single RGB image $I\in \mathbb{R}^{M\times 3}$ with $M$ pixels as input and predicts a 3D semantic occupancy volume $\hat{Y}\in \mathbb{R}^{N}$ of a scene, where $N$ denotes the number of scene voxels. Each voxel $y_i$ in $\hat{Y}$ is either empty or occupied with one semantic class. A 3D occupancy model typically consists of an occupancy network $\mathcal{N}$ and an occupancy head $\mathcal{H}$ as follows:
\begin{equation}
V_{3D}=\mathcal{N}\left(I\right), \hat{Y}=\mathcal{H}\left(V_{3D}\right),
\end{equation}
where the network $\mathcal{N}$ generates 3D voxel features with $D$ dimensions $V_{3D}\in \mathbb{R}^{N\times D}$ from 2D image input $I$, and the decoder $\mathcal{H}$ produces the semantic occupancy prediction $\hat{Y}$.

Existing methods train occupancy models with ground-truth occupancy semantics as direct supervision. We instead propose a self-supervised pre-training paradigm that learns powerful 3D voxel representations via distillation from vision foundation models, avoiding manual annotations. Our approach requires only minor architectural changes to the occupancy network: the sole difference between pre-training and fine-tuning is the dimensionality $D$ of the feature $V_{3D}$. Thus, the framework is model-agnostic, and we evaluate it on three distinct occupancy networks in Sec.~\ref{sec:exp}.

\subsection{Pre-Training Stage}
\label{sec:pre-training-stage}
The goal of our pre-training framework is to learn dense 3D representations by distilling knowledge from powerful 2D foundation models. As illustrated in Fig.~\ref{fig:pre-training}, our framework is built upon two complementary alignment objectives: an intra-frame feature alignment for local semantic coherence, and a cross-scene feature alignment for global semantic consistency.

\textbf{Overview.} Building on the occupancy network $\mathcal{N}$, we construct the 3D voxel representation $V_{3D}$. For an image $I$, we first obtain a semantic region map $\mathcal{M}_{sem}$ via our pixel-to-text association module. This map groups features from two heterogeneous streams: a 2D teacher and a 3D student. The 2D teacher stream extracts dense DINOv2 features $\mathcal{F}_{2D}$ and aggregates them with $\mathcal{M}_{sem}$ to form region-level embeddings $R_{2D}$. The 3D student stream projects 3D voxel features into a 2D feature map $\hat{\mathcal{F}}_{2D}$ (via voxel-anchored Gaussianization), then uses the same $\mathcal{M}_{sem}$ to obtain rendered region-level embeddings $R_{3D}$. Given $R_{2D}$ and $R_{3D}$, we perform intra-frame alignment with a region contrastive loss $\mathcal{L}_{region}$, which aligns $R_{3D}$ to the corresponding $R_{2D}$ in each frame. To  enforce cross-scene feature alignment, we propose a semantic prototype update module that utilizes $\mathcal{F}_{2D}$ and $\mathcal{M}_{sem}$ to update class prototypes $\mathcal{P}$, thereby achieving a prototype contrastive loss $\mathcal{L}_{proto}$ that pulls $R_{3D}$ toward their corresponding semantic prototypes in $\mathcal{P}$.

\subsubsection{Intra-Frame Feature Alignment}
This module aligns region-level embeddings between 3D and 2D modalities within a single frame, enhancing robustness against artifacts and boundary noise inherent in foundation models.

\textbf{(a) 2D Teacher Feature Generation.} 
The teacher stream produces distillation targets by integrating dense 2D features $\mathcal{F}_{2D}$ from DINOv2 with semantic masks $\mathcal{M}_{sem}$. As in Sec.~\ref{sec:data_pipeline}, an open-vocabulary semantic model generates $C$ binary masks based on pre-defined class prompts. The target region feature $R_{2D}^c$ for each class $c$ is then computed via masked average pooling over $\mathcal{F}_{2D}$.

\textbf{(b) Voxel-Anchored Gaussianization.} 
To bridge the modality gap, we transform the 3D voxel feature $V_{3D}$ into 3D Gaussian primitives for differentiable rendering. Each Gaussian primitive is parameterized as follows: (1) \textit{Position} is derived from the voxel coordinates; (2) \textit{Features} are inherited from $V_{3D}$; (3) \textit{Opacity} is predicted via an MLP; (4) \textit{Scale and Rotation} are fixed to the voxel size and identity matrix, respectively. We then rasterize these primitives to produce a rendered 2D feature map $\hat{F}_{2D}$. Finally, the rendered region embedding $R_{3D}^c$ is obtained by applying $\mathcal{M}_{sem}$ to $\hat{F}_{2D}$ through masked average pooling.

\textbf{(c) Region Contrastive Loss.} Given the $R_{3D}$ and the $R_{2D}$, we  derive the region-level loss $\mathcal{L}_{region}$:
\begin{equation}
\mathcal{L}_{region} = -\sum_{i=1}^C \log \frac{\exp(\langle f_{3D}^i, f_{2D}^i \rangle / \tau_{region})}{\sum_{j=1}^{C} \exp(\langle f_{3D}^i, f_{2D}^{j} \rangle / \tau_{region})},
\end{equation}
where $f_{3D}^i$/$f_{2D}^i$ is the $i^{th}$ region feature embedding from $R_{3D}$/$R_{2D}$, $\langle \cdot, \cdot \rangle$ denotes scalar product, and $\tau_{region}$ is a temperature hyper-parameter. This enables the 3D occupancy network to learn representations aligned with the spatial and semantic understanding of the powerful 2D foundation model.

\subsubsection{Cross-Scene Feature Alignment.}
While intra-frame alignment ensures local coherence, a robust 3D representation must also exhibit global semantic consistency across diverse scenes. To achieve this, we introduce a cross-scene feature alignment mechanism based on semantic prototype contrastive learning. This module is depicted in the upper part of Fig.~\ref{fig:pre-training}.

\textbf{(a) Prototype Generation.} We maintain a dynamic memory bank of class-specific prototypes $\mathcal{P} = \{p_1, p_2, ..., p_C\}$, where $C$ is the number of semantic classes. Each prototype $p_c$ is a feature vector that represents the centroid of a particular semantic class in the embedding space. This memory bank is updated online during training. For each training sample, we collect  $\mathcal{F}_{2D}$ and $\mathcal{M}_{sem}$ produced during the 2D teacher feature generation step. These features are then used to update their corresponding class prototypes in the memory bank, typically via an exponential moving average with momentum $\beta$. This update rule allows the prototypes to evolve smoothly, capturing the canonical representation of each class from diverse samples observed throughout the entire dataset.

\textbf{(b) Prototype Contrastive Loss.} We regularize the $R_{3D}$ by contrasting them against the global class prototypes $\mathcal{P}$. We define the prototype contrastive loss as $\mathcal{L}_{proto}$:
\begin{equation}
\mathcal{L}_{proto}=-\sum_{i=1}^{C}\log\frac{\exp(\langle f_{3D}^{i},p_{i}\rangle/\tau_{proto})}{\sum_{j=1}^{C}\exp(\langle f_{3D}^{i},p_{j}\rangle/\tau_{proto})}.
\end{equation} 
By minimizing this loss, we enforce a globally consistent and well-structured feature space, compelling the 3D semantic occupancy model to learn representations that are truly discriminative and generalizable across diverse scenes. Ultimately, our pre-training loss is given by:
\begin{equation}
\mathcal{L}_{\text{total}} = \mathcal{L}_{region} + \mathcal{L}_{proto}.
\label{eq:total_loss}
\end{equation}

\subsection{Discussion}
\label{sec:pre-training-discussion}
Similar to recent methods in self-supervised outdoor semantic occupancy (\emph{e.g.}, GaussTR \cite{jiang2025gausstr}, AGO \cite{li2025ago}, AutoOcc \cite{zhou2025autoocc}, SelfOcc \cite{huang2024selfocc}, and S4C \cite{hayler2023s4c}), our framework leverages 2D and 3D foundation models. However, our approach markedly diverges from these excellent works in the following two aspects: (1) Robustness to Uncalibrated Video: These outdoor methods rely on the assumption that either exact camera setups or predefined vehicle platforms are available, and their evaluation is conducted almost exclusively within the same data domain. Transferring these approaches to unconstrained, in-the-wild videos devoid of camera priors is nearly infeasible. (2) Model-Agnostic Pre-training: Rather than developing specialized, tightly coupled architectures, we introduce a model-agnostic pre-training framework that could enhance diverse indoor occupancy networks.

\section{Experiments}
\label{sec:exp}

The experimental setups are presented in Sec. \hyperref[sec:experiment_settings]{5.1}. Sec. \hyperref[sec:evaluation_of_data_pipeline]{5.2} evaluates the data pipeline using various methods, and Sec. \hyperref[sec:evaluation_of_pretraining]{5.3} evaluates the pre-training paradigm on different architectures and datasets. Sec. \hyperref[sec:ablation_study]{5.4} provides ablation studies to assess each component.

\subsection{Experiment Settings}
\label{sec:experiment_settings}

\textbf{Datasets.}\quad We evaluate on two standard 3D semantic occupancy benchmarks, NYUv2 \cite{silberman2012indoor} and Occ-ScanNet \cite{yu2024monocular}, alongside our proposed YouTube-Occ. All datasets share a $60\times60\times36$ occupancy grid at 8 cm resolution, annotated with 13 classes (11 semantic, 1 free, 1 unknown). NYUv2 and Occ-ScanNet contain 795/654 and 45,755/19,764 train/test images, respectively, while YouTube-Occ contributes 100,372 frames across 5,241 scenes. Following \cite{cao2022monoscene}, we report IoU and mIoU for all semantic classes.

\textbf{Pre-training and Fine-tuning Details.} We validate our method across three diverse architectures: MonoScene (Voxel-based CNN) \cite{cao2022monoscene}, Symphonies (Voxel-based Transformer) \cite{jiang2024symphonize}, and EmbodiedOcc (Gaussian-based Transformer) \cite{wu2024embodiedocc}. Pre-training utilizes DINOv2 \cite{oquab2023dinov2} for 2D features and SAN \cite{xu2023side} for text-to-pixel association. During fine-tuning, the occupancy network retains pre-trained weights while the head is trained from scratch. Both stages employ the AdamW optimizer (initial lr=2e-4) with a 0.1 multi-step decay at predefined epochs. Following \cite{yu2024monocular}, we train for 30 epochs on NYUv2 and 10 on Occ-ScanNet and YouTube-Occ. Experiments are conducted on four RTX A6000 GPUs with a batch size of 4.

\begin{table}[htbp]
\centering
\caption{Evaluation of pseudo labels with various methods on Occ-ScanNet. Symphonies serves as the base model, and pseudo-labels are generated by our data pipeline.}
\label{tab:zeroshot_occscannet}

\resizebox{0.6\linewidth}{!}{
\begin{tabular}{@{}l@{\hspace{2em}}lcc@{}}
\toprule
\textbf{Method} & \textbf{Training Data} & \textbf{mIoU$\uparrow$} & \textbf{IoU$\uparrow$} \\ 
\midrule
Pseudo-Label    & None                      & 10.06 & 27.00 \\
Zero-Shot       & YouTube-Occ              & 8.41  & 12.74 \\
Joint-Training  & NYUv2 + Occ-ScanNet       & 40.17 & 51.97 \\
Joint-Training  & YouTube-Occ + Occ-ScanNet & 37.38 & 55.06 \\
Fine-Tuning     & Occ-ScanNet               & \textbf{47.71} & \textbf{59.89} \\
\bottomrule
\end{tabular}
}
\end{table}

\subsection{Evaluation of Data Pipeline}
\label{sec:evaluation_of_data_pipeline}
\begin{figure}[htbp]
    \centering
    \includegraphics[width=0.75\textwidth]{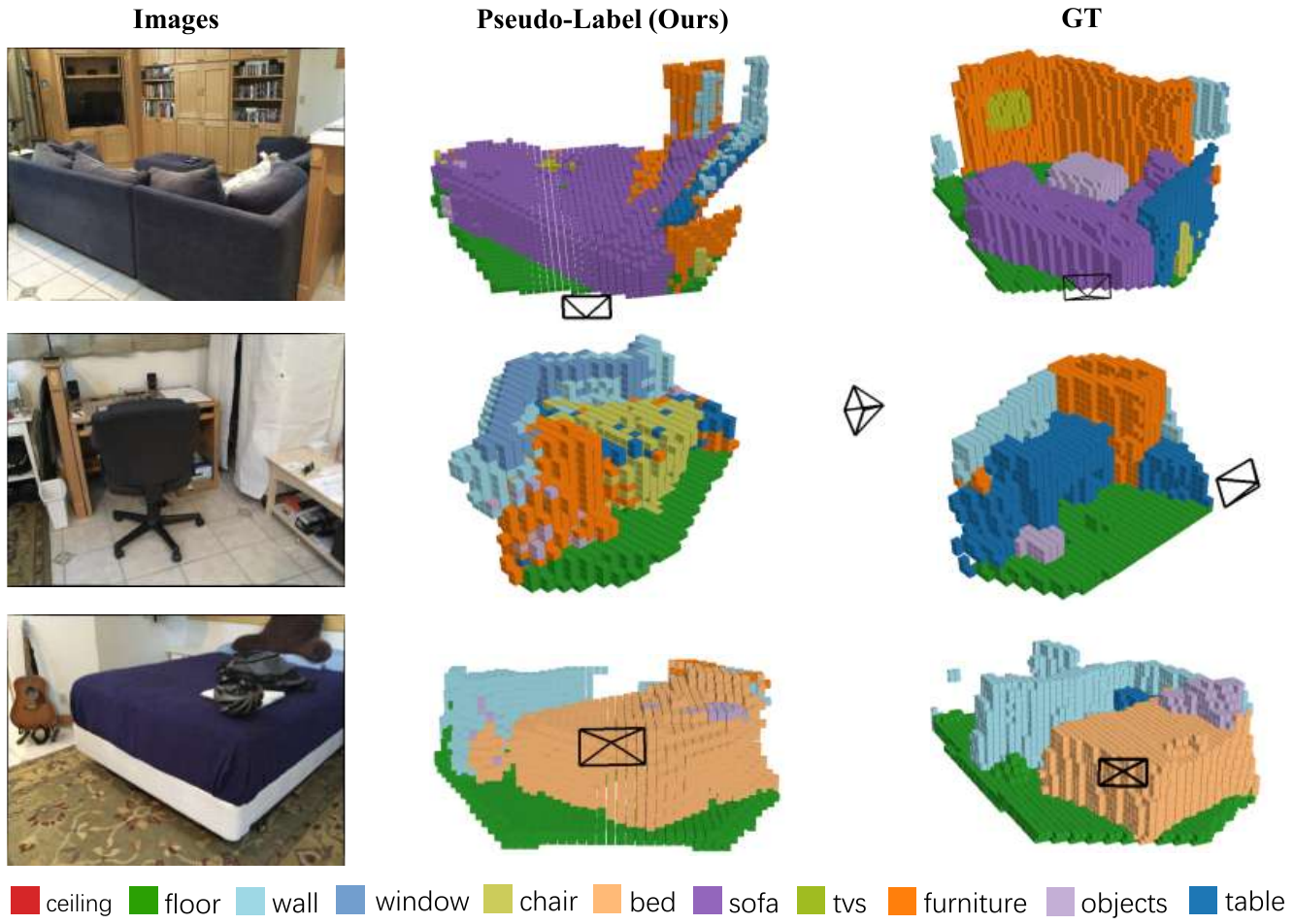}
    \caption{ 3D Semantic Occupancy Pseudo-label visualization in Occ-ScanNet}
    \label{fig:pseudo_label_vis}
\end{figure}

Prior to utilizing unconstrained internet data, we first evaluate the quality of the pseudo-labels from our pipeline on Occ-ScanNet, which provides ground-truth 3D semantic occupancy annotations. Specifically, we adopt common supervised learning strategies (see Tab.~\ref{tab:zeroshot_occscannet}), and present qualitative results in Fig.~\ref{fig:pseudo_label_vis}. Directly training Symphonies on YouTube-Occ and validating on Occ-ScanNet (Zero-Shot) yields an unsatisfactory result of 8.41\% mIoU, which is even lower than evaluating the raw pseudo-labels (10.06\% mIoU). Furthermore, naively jointly training Occ-ScanNet and YouTube-Occ fails to bring consistent improvements, degrading the mIoU compared to the NYUv2 joint-training baseline (37.38\% vs. 40.17\%). These results indicate that simply applying standard supervision is insufficient to harness the potential of these internet-derived pseudo-labels, necessitating a more sophisticated approach. To this end, we develop a pre-training framework to exploit the potential of our collected in-the-wild videos.

\subsection{Evaluation of Pre-training Framework}
\label{sec:evaluation_of_pretraining}
Due to the absence of prior work that leverages web data for semantic occupancy, we evaluate the pre-training framework from three perspectives: (1) existing models with/without pre-training, (2) impact of pre-training on web data, and (3) comparison with alternative plug-and-play methods.
\vspace{-1em}

\begin{table}[htbp]
\caption{Performance comparison on NYUv2 and Occ-ScanNet. All methods are evaluated using the standard IoU and mIoU metrics. \textbf{Bold} denotes the highest score. \textdagger indicates reproduced results using the official public code.}
\centering
\resizebox{0.9\linewidth}{!}{

\setlength{\tabcolsep}{3pt} 
\small 

\begin{tabular}{ 
  l |         
  *{11}{c} |  
  cc          
}

\textbf{Method}  
& \rotatebox{90}{\parbox{1.5cm}{\textcolor{ceiling}{$\blacksquare$} ceiling}} 
				& \rotatebox{90}{\textcolor{floor}{$\blacksquare$} floor}
				& \rotatebox{90}{\textcolor{wall}{$\blacksquare$} wall} 
				& \rotatebox{90}{\textcolor{window}{$\blacksquare$} window} 
				& \rotatebox{90}{\textcolor{chair}{$\blacksquare$} chair} 
				& \rotatebox{90}{\textcolor{bed}{$\blacksquare$} bed} 
				& \rotatebox{90}{\textcolor{sofa}{$\blacksquare$} sofa} 
				& \rotatebox{90}{\textcolor{table}{$\blacksquare$} table} 
				& \rotatebox{90}{\textcolor{tvs}{$\blacksquare$} tvs} 
				& \rotatebox{90}{\textcolor{furniture}{$\blacksquare$} furniture} 
				& \rotatebox{90}{\textcolor{objects}{$\blacksquare$} objects}

& {\textbf{mIoU$\uparrow$}}  & {\textbf{IoU$\uparrow$}}

\\
\midrule

\multicolumn{14}{>{\columncolor{sectionblue}}l}{\textcolor{sectiontext}{\textit{\textbf{\hspace{0.5em} NYUv2}}}} \\
LMSCNet \cite{roldao2020lmscnet}     & 4.49  & 88.41 & 4.63  & 0.25  & 3.94  & 32.03 & 15.44 & 6.57  & 0.02  & 14.51 & 4.39  & 15.88 & 33.93 \\
AICNet \cite{li2020anisotropic}      & 7.58  & 82.97 & 9.15  & 0.05  & 6.93  & 35.87 & 22.92 & 11.11 & 0.71  & 15.90 & 6.45  & 18.15 & 30.03 \\
3DSketch \cite{chen20203d}     & 8.53  & 90.45 & 9.94  & 5.67  & 10.64 & 42.29 & 29.21 & 13.88 & 9.38  & 23.83 & 8.19  & 22.91 & 38.64 \\
MonoScene \cite{cao2022monoscene}   & 8.89  & 93.50 & 12.06 & 12.57 & 13.72 & 48.19 & 36.11 & 15.13 & 15.22 & 27.96 & 12.94 & 26.94  & 42.51 \\
NDC-Scene \cite{yao2023ndc}  & 12.02 &  93.51 & 13.11 & 13.77 & 15.83 &  49.57 &  39.87 & 17.17 &  24.57 & 31.00 & 14.96 & 29.03  & 44.17 \\
ISO \cite{yu2024monocular}  & 14.21 &  93.47 & 15.89 & 15.14 &\bfseries 18.55 &\bfseries 50.01 & 40.82 &\bfseries 18.25 & \bfseries25.90 & 34.08 & 17.67 & 31.25  & 47.11 \\
\arrayrulecolor{gray}
\midrule
\arrayrulecolor{black}
MonoScene\textdagger  & 8.52 & 93.50 & 11.67 & 10.81 & 12.87 & 47.23 & 35.81 & 14.82 & 12.56 & 26.06 & 13.17 & 26.09 & 41.87 \\
\rowcolor{rankonegreen}
\hspace{1em} w/ Pre-training  & 8.73 &\bfseries 93.63 & 12.38 & 11.66 & 13.77 & 48.13 & 37.09 & 15.67 & 14.66 & 27.91 & 13.62 & 27.02 & 42.61 \\

EmbodiedOcc\textdagger  & 3.50 & 67.20 & 2.30 & 6.90 & 7.10 & 35.60 & 33.10 & 11.50 & 17.00 & 18.30 & 8.10 & 19.14 & 30.46 \\
\rowcolor{rankonegreen}
\hspace{1em} w/ Pre-training  & 5.20 & 70.00 & 6.10 & 7.30 & 8.00 & 34.90 & 32.90 & 12.10 & 17.90 & 18.20 & 9.30 & 20.18 & 31.85\\
Symphonies \textdagger  & 14.54 & 86.59 & 25.95 & 15.69 & 16.78 & 46.60 & 38.06 & 15.37 & 15.33 & 32.16 & 19.58 & 29.70 & 49.91\\
\rowcolor{rankonegreen}
\hspace{1em} w/ Pre-training  &\bfseries 14.98 & 93.13 &\bfseries 26.54 & \bfseries18.36 & 16.88 & 48.15 & \bfseries41.37 & 17.62 & 15.80 & \bfseries34.72 &\bfseries 20.64 & \bfseries31.65 & \bfseries52.15\\

\addlinespace[1pt] 
\bottomrule
\multicolumn{14}{>{\columncolor{sectionblue}}l}{\textcolor{sectiontext}{\textit{\textbf{\hspace{0.5em} Occ-ScanNet}}}} \\
MonoScene \cite{cao2022monoscene}      & 15.17 & 44.71 & 22.41 & 12.55 & 26.11 & 27.03 & 35.91 & 28.32 & 6.57  & 32.16 & 19.84 & 24.62 & 41.60 \\
ISO \cite{yu2024monocular}          & 19.88 & 41.88 & 22.37 & 16.98 & 29.09 & 42.43 & 42.00 & 29.60 & 10.62 & 36.36 & 24.61 & 28.71 & 42.16 \\
EmbodiedOcc \cite{wu2024embodiedocc}   & 40.90 & 50.80 & 41.90 & 33.00 & 41.20 & 55.20 &  61.90 & 43.80 & 35.40 &  53.50 & 42.90 & 45.48 & 53.95 \\
EmbodiedOcc\texttt{++} \cite{wang2025embodiedocc++}  & 36.40 & 53.10 & 41.80 & 34.40 & 42.90 & 57.30 & \bfseries 64.10 & 45.20 & 34.80 & 54.20 & 44.10 & 46.20 & 54.90 \\
RoboOcc \cite{zhang2025roboocc}    & 45.36 & 53.49 & 44.35 & 34.81 & 43.38 & 56.93 & 63.35 & 46.35 & 36.12 & \bfseries 55.48 & 44.78 & 47.67 & 56.48 \\
\arrayrulecolor{gray}
\midrule
\arrayrulecolor{black}
EmbodiedOcc\textdagger   & 40.30 & 50.30 & 41.60 & 32.10 & 40.80 & 52.70 & 60.00 & 43.60 & 36.00 & 52.30 & 42.00 & 44.69 & 53.23 \\

\rowcolor{rankonegreen}

\hspace{1em} w/ Pre-training   & 42.20 & 50.00 & 41.70 & 34.60 & 41.40 & 57.20 & 61.70 & 44.20 & 38.50 & 53.70 & 42.70 & 46.17 & 53.80 \\
Symphonies\textdagger   & 42.14 &  57.33 & 47.92 & 38.50 & 43.58 & 56.99 & 60.16 & 45.42 & 37.44 & 51.54 & 43.80 & 47.71  & 59.89 \\
\rowcolor{rankonegreen}
\hspace{1em} w/ Pre-training     & \bfseries 46.81 &\bfseries 57.85 & \bfseries 50.29 & \bfseries 40.47 & \bfseries 44.75 & \bfseries 57.36 & 60.33 & \bfseries 47.00 & \bfseries 38.88 & 53.26 & \bfseries 46.21 & \bfseries 49.38 & \bfseries 61.25 \\
\addlinespace[1pt]
\bottomrule
\end{tabular}
}

\label{tab:table_100}
\vspace{-2em}
\end{table}

\subsubsection{Existing Models Equipped with Pre-Training}
\begin{figure}[htbp]
\centering
\vspace{-1em}
\includegraphics[width=0.90\linewidth]{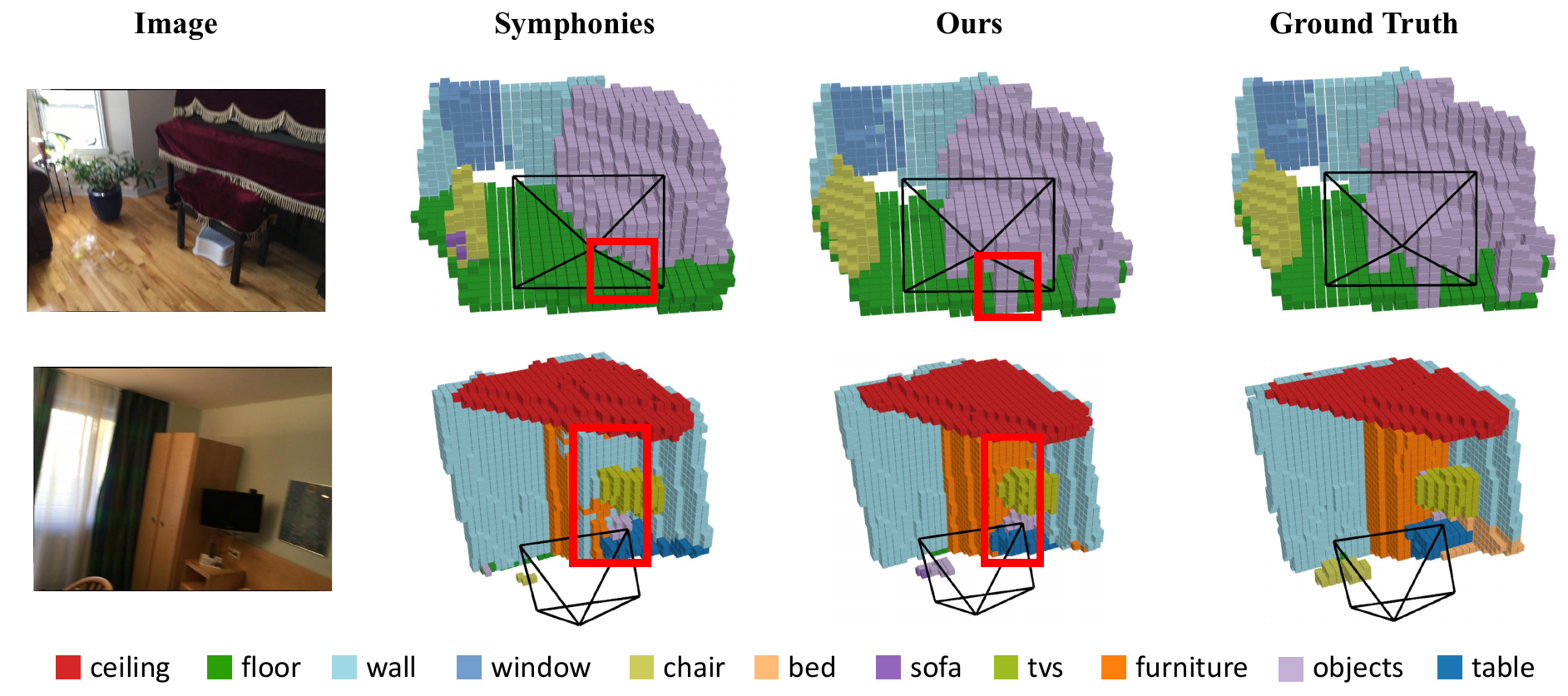}
\caption{ Qualitative results of the model with pre-training on the Occ-ScanNet dataset.}
\label{fig:vis_occscannet}
\vspace{-2em}
\end{figure}

Tab.~\ref{tab:table_100} demonstrates that our pre-training framework consistently enhances performance across multiple architectures and benchmarks. Integrated with Symphonies~\cite{jiang2024symphonize}, our method achieves state-of-the-art (SOTA) results, reaching 31.65\% mIoU (+1.95\%) on NYUv2 and 49.38\% mIoU (+1.67\%) on Occ-ScanNet. Crucially, these improvements are model-agnostic. We observe consistent gains across diverse baselines, including EmbodiedOcc (+1.04\% and +1.48\% on NYUv2 and Occ-ScanNet, respectively) and MonoScene (+0.93\% on NYUv2). This indicates that our framework learns robust and generalizable 3D priors. Qualitative results (Fig.~\ref{fig:vis_occscannet}) further confirm that our approach yields more structurally coherent and accurate semantic completions than the baselines.

\subsubsection{Pre-training on YouTube-Occ}
Beyond fine-tuning on the full dataset, we further conduct a few-shot evaluation to assess the quality of representations learned from YouTube-Occ under limited supervision (Tab.~\ref{tab:table_internet_fewshot}). We pre-train the Symphonies model on two scales of our dataset, the full YouTube-Occ and a 50K-sample subset (YouTube-Occ-50K), and compare them against random initialization and in-domain pre-training baselines.

Evaluations demonstrate that YouTube-Occ  enhances data-efficient learning. On NYUv2, our pre-training provides a superior initialization compared to random weights, achieving 18.81\% mIoU (+3.29\%) with only 5\% of the labels, and a +5.09\% gain at the 10\% split. These benefits extend to the Occ-ScanNet benchmark, where pre-training delivers a +2.36\% mIoU boost in the 10\% few-shot setting. Furthermore, we observe a clear scaling trend: the full YouTube-Occ dataset consistently outperforms the 50K subset. Overall, these results confirm that the scale and diversity of our dataset equip 3D occupancy models with robust, generalizable semantic representations, improving both downstream performance and data efficiency.

\vspace{-1em}
\begin{table}[htbp]
\caption{Impact of pre-training on few-shot 3D semantic occupancy prediction. We evaluate Symphonies pre-trained on different data sources, fine-tuned with varying labeled data ratios on NYUv2 and Occ-ScanNet. \textbf{Bold} numbers mark the best result for each setting, and \textcolor{green}{green} numbers show the gain over random initialization.} 
\centering
\resizebox{1.0\linewidth}{!}{
\small 
\begin{tabular}{
l|cc|cc|cc|cc|cc
}
\toprule
\textbf{Pre-training} & \multicolumn{2}{c|}{\textbf{5\%}} & \multicolumn{2}{c|}{\textbf{10\%}} & \multicolumn{2}{c|}{\textbf{20\%}} & \multicolumn{2}{c|}{\textbf{50\%}} & \multicolumn{2}{c}{\textbf{Full}} \\
\cmidrule(lr){2-3} \cmidrule(lr){4-5} \cmidrule(lr){6-7} \cmidrule(lr){8-9} \cmidrule(lr){10-11}
\textbf{DataSource} & \textbf{mIoU$\uparrow$} & \textbf{IoU$\uparrow$} & \textbf{mIoU$\uparrow$} & \textbf{IoU$\uparrow$} & \textbf{mIoU$\uparrow$} & \textbf{IoU$\uparrow$} & \textbf{mIoU$\uparrow$} & \textbf{IoU$\uparrow$} & \textbf{mIoU$\uparrow$} & \textbf{IoU$\uparrow$} \\
\midrule
\multicolumn{11}{>{\columncolor{sectionblue}}l}{\textcolor{sectiontext}{\textit{\textbf{\hspace{0.5em} NYUv2}}}} \\
Random       & 15.52 & 42.61 & 17.43 & 43.99 & 23.26 & 46.49 & 29.15 & 49.71 & 29.70 & 49.91 \\
YouTube-Occ-50K  & 18.25 & 43.19 & 22.36 & 44.58 & 25.20 & 46.82 & 29.51 & 49.77 & 30.87 & 51.37 \\
(gain) & \textcolor{green}{+2.73} & \textcolor{green}{+0.58} & \textcolor{green}{+4.93} & \textcolor{green}{+0.59} & \textcolor{green}{+1.94} & \textcolor{green}{+0.33} & \textcolor{green}{+0.36} & \textcolor{green}{+0.06} & \textcolor{green}{+1.17} & \textcolor{green}{+1.46} \\

YouTube-Occ & 18.81 & 44.04 & 22.52 & 45.85 & 26.01 & 47.43 & 29.97 & 50.36 & 31.45 & \bf 52.88 \\
(gain) & \textcolor{green}{+3.29} & \textcolor{green}{+1.43} & \textcolor{green}{+5.09} & \textcolor{green}{+1.86} & \textcolor{green}{+2.75} & \textcolor{green}{+0.94} & \textcolor{green}{+0.82} & \textcolor{green}{+0.65} & \textcolor{green}{+1.75} & \textcolor{green}{+2.97} \\
NYUv2          & \bf 19.87 &\bf 44.23 &\bf 23.58 &\bf 46.37 &\bf 26.64 &\bf 48.37 &\bf 30.30 &\bf 50.83 &\bf 31.65 & 52.15 \\
(gain) & \textcolor{green}{+4.35} & \textcolor{green}{+1.62} & \textcolor{green}{+6.15} & \textcolor{green}{+2.38} & \textcolor{green}{+3.38} & \textcolor{green}{+1.88} & \textcolor{green}{+1.15} & \textcolor{green}{+1.12} & \textcolor{green}{+1.95} & \textcolor{green}{+2.24} \\

\midrule
\multicolumn{11}{>{\columncolor{sectionblue}}l}{\textcolor{sectiontext}{\textit{\textbf{\hspace{0.5em} Occ-ScanNet}}}} \\
Random       & 31.77 & 49.43 & 34.18 & 50.16 & 38.06 & 53.26 & 43.74 & 57.58 & 47.71 & 59.89 \\
YouTube-Occ-50K  & 33.18 & 49.92 & 36.26 & 51.59 & 38.90 &\bf 54.19 & 44.34 & 58.03 & 48.15 & 60.33 \\
(gain) & \textcolor{green}{+1.41} & \textcolor{green}{+0.49} & \textcolor{green}{+2.08} & \textcolor{green}{+1.43} & \textcolor{green}{+0.84} & \textcolor{green}{+0.93} & \textcolor{green}{+0.60} & \textcolor{green}{+0.45} & \textcolor{green}{+0.44} & \textcolor{green}{+0.43} \\
YouTube-Occ & 33.50 &\bf 50.42 & 36.54 & 51.78 & 39.33 & 53.57 & 44.60 & 57.90 & 48.47 & 60.63 \\
(gain) & \textcolor{green}{+1.73} & \textcolor{green}{+0.99} & \textcolor{green}{+2.36} & \textcolor{green}{+1.62} & \textcolor{green}{+1.27} & \textcolor{green}{+0.31} & \textcolor{green}{+0.86} & \textcolor{green}{+0.32} & \textcolor{green}{+0.76} & \textcolor{green}{+0.73} \\
Occ-ScanNet  &\bf 33.87 & 49.60 &\bf 37.80 &\bf 52.56 &\bf 39.63 & 53.94 &\bf 45.86 &\bf 58.59 &\bf 49.38 &\bf 61.25 \\
(gain) & \textcolor{green}{+2.10} & \textcolor{green}{+0.17} & \textcolor{green}{+3.62} & \textcolor{green}{+2.40} & \textcolor{green}{+1.57} & \textcolor{green}{+0.68} & \textcolor{green}{+2.12} & \textcolor{green}{+1.01} & \textcolor{green}{+1.67} & \textcolor{green}{+1.35} \\

\bottomrule
\end{tabular}
}

\label{tab:table_internet_fewshot}
\vspace{-2em}
\end{table}

\subsubsection{Comparison to Existing Strategies}
We compare our pre-training framework against other training strategies in Tab.~\ref{tab:training_strategy}. Our approach outperforms methods that rely on auxiliary losses, specifically HASSC~\cite{wang2024hassc} and GaussRender~\cite{chambon2025gaussrender}. Moreover, previous 3D pre-training method Pri3D~\cite{Hou_2021_ICCV} even underperforms the vanilla baseline, likely because it only enforces pixel-voxel consistency with an isolated model, limiting transferability. These results demonstrate the effectiveness of our pre-training approach for improving 3D semantic occupancy estimation.

\begin{table}[htbp]
\vspace{-1em}
\caption{Comparison of different training strategies on the NYUv2 dataset.
Results are reported in terms of mIoU/IoU (\%).}
\centering
\scalebox{0.95}{
\begin{tabular}{lccc}
\toprule
\textbf{Method}
& \textbf{Training Strategy}
& \multicolumn{2}{c}{\textbf{Occupancy Networks}} \\
\cmidrule(lr){3-4}
& & \textbf{Symphonies} & \textbf{EmbodiedOcc} \\
\midrule
Vanilla
& Standard
& 29.70/49.91
& 19.14/30.46 \\

w/ HASSC~\cite{wang2024hassc}
& Auxiliary Loss
& 30.70/51.15
& 19.54/30.42 \\

w/ GaussRender~\cite{chambon2025gaussrender}
& Auxiliary Loss
& 30.37/50.77
& 19.22/30.96 \\

w/ Pri3D~\cite{Hou_2021_ICCV}
& Pre-training
& 28.40/49.32
& 17.94/29.01 \\

w/ Pre-training (Ours)
& Pre-training
& \textbf{31.65/52.15}
& \textbf{20.18/31.85} \\
\bottomrule
\end{tabular}%
}
\label{tab:training_strategy}
\vspace{-3em}
\end{table}

\subsection{Ablation Study}
\label{sec:ablation_study}
To investigate the individual contributions of our proposed modules, Tab.~\ref{tab:pretrain_ablation_exp} ablates the impact of the two loss components, $\mathcal{L}_{region}$ and $\mathcal{L}_{proto}$. The baseline model yields 29.70\% and 47.71\% mIoU on NYUv2 and Occ-ScanNet, respectively. Incorporating $\mathcal{L}_{region}$ for intra-frame alignment improves the NYUv2 mIoU to 31.12\%, while introducing $\mathcal{L}_{proto}$ for cross-scene consistency achieves 31.24\%. Combining both components yields the best performance, reaching 31.65\% on NYUv2 and 49.38\% on Occ-ScanNet. These correspond to absolute gains of +1.95\% and +1.67\% over the baseline, demonstrating that local region alignment and global prototype consistency are highly complementary for 3D occupancy representation learning.

\begin{table}[htbp]
\vspace{-1em}
\caption{Ablation study of the two feature-alignment losses in our pre-training framework.}
\centering
\scalebox{1.0}{
\begin{tabular}{cccccc}
\toprule
\multirow{2}{*}{\textbf{Region Loss}} & \multirow{2}{*}{\textbf{Prototype Loss}} & \multicolumn{2}{c}{\textbf{NYUv2}} & \multicolumn{2}{c}{\textbf{Occ-ScanNet}} \\
\cmidrule(lr){3-4} \cmidrule(lr){5-6}
 & & \bf mIoU$\uparrow$ &\bf IoU$\uparrow$ &\bf mIoU$\uparrow$ &\bf IoU$\uparrow$ \\
\midrule
& & 29.70 & 49.91 & 47.71 & 59.89 \\
\checkmark & &  31.12 & 52.09 & 48.15 & 60.31 \\
& \checkmark & 31.24 & 51.76 & 48.76 & 60.47 \\
\checkmark & \checkmark & \bf 31.65 & \bf 52.15 & \bf 49.38 & \bf  61.25 \\
\bottomrule
\end{tabular}
}

\label{tab:pretrain_ablation_exp}
\vspace{-2em}
\end{table}
Tab.~\ref{tab:ablation_study_gaussian_and_prototype} ablates the proposed projection and prototype mechanisms on the NYUv2 dataset. The default configuration, utilizing Gaussianization and a momentum of $\beta = 0.999$, yields the best performance at 31.65\% mIoU and 52.15\% IoU. Replacing Gaussianization with direct 3D projection or introducing learnable scale and rotation leads to notable performance drops of $-1.07\%$ and $-1.76\%$ mIoU, respectively, justifying the effectiveness of the introduced geometric prior. 
Similarly, altering the prototype momentum to $0.9$ or applying a semantic threshold of $0.3$ degrades performance. This could indicate that our prototype-based module can alleviate the inevitable noise in pseudo labels.

\begin{table}[htbp]
\vspace{-1em}
\caption{Ablation study of the projection and prototype mechanisms on the NYUv2 dataset.}
\centering
\scalebox{1.0}{
\begin{tabular}{@{}lcc@{}}
\toprule
\textbf{Configuration} & \textbf{mIoU$\uparrow$} & \textbf{IoU$\uparrow$} \\
\midrule
Default (Gaussianization and $\beta=0.999$) & \textbf{31.65} & \textbf{52.15} \\
\midrule
\multicolumn{3}{@{}l}{\textit{Projection Variants}} \\
Direct 3D Projection & 30.58 & 50.83 \\
Learnable Scale \& Rotation & 29.89 & 50.53 \\
\midrule
\multicolumn{3}{@{}l}{\textit{Prototype Variants}} \\
Momentum $\beta = 0.9$ & 30.37 & 50.34 \\
Semantic Threshold $= 0.3$ & 30.42 & 50.30 \\
\bottomrule
\end{tabular}
}

\label{tab:ablation_study_gaussian_and_prototype}
\vspace{-3em}
\end{table}

\section{Conclusion and Limitation}
Our work introduces YouTube-Occ, an automated pipeline for generating pseudo-labeled 3D data from uncalibrated internet videos, alongside a self-supervised distillation framework that boosts 3D occupancy networks by leveraging foundation models. Experimental results on NYUv2 and Occ-ScanNet validate that our data pipeline and pre-training framework deliver consistent improvements, particularly in data-scarce settings. This study demonstrates that rich 3D priors can be extracted from in-the-wild videos, underscoring web-scale data as a promising resource for 3D perception. \textbf{Limitations} of our method include handling invisible or occluded voxels, extending to multi-view or temporal networks, and generalizing beyond the wall height assumption. 

\vspace{-1em}
\section*{Acknowledgements}
This work is supported by the National Natural Science Foundation of China (Grant No. 62502159, W2521174, 62476090, U23A20343, 62302167, 62222602), Fundamental Research Funds for the Central Universities, Natural Science Foundation of Shanghai (Grant No. 25ZR1402135),  the Shanghai Committee of Science and Technology (Grant No. 25511103300, 25511104302, 25511102700), ``Chen Guang'' project supported by Shanghai Municipal Education Commission and Shanghai Education Development Foundation (Grant No. 25CGA22), Natural Science Foundation of Chongqing (Grant No. CSTB2025NSCQ-GPX0445), Young Elite Scientists Sponsorship Program by CAST (Grant No. YESS20240780), Open Project Program of the State Key Laboratory of CAD\&CG (Grant No. A2501), Zhejiang University, Open Research Fund of Key Laboratory of Advanced Theory and Application in Statistics and Data Science-MOE, ECNU.

%
%
\bibliographystyle{splncs04}
\bibliography{main}


\end{document}